\documentclass[10pt,twocolumn,letterpaper]{article}

\usepackage{cvpr}
\usepackage{times}
\usepackage{epsfig}
\usepackage{graphicx}
\usepackage{amsmath}
\usepackage{amssymb}
\usepackage{graphicx}
\usepackage{algorithm}
\usepackage{algorithmic}
\usepackage{wrapfig}
\usepackage{multirow}
\usepackage{bigdelim}
\usepackage{microtype}
\usepackage{amsmath}
\usepackage{amssymb}
\usepackage{latexsym}
\usepackage{enumitem}
\usepackage{url}
\usepackage{xcolor}
\usepackage{dblfloatfix}
\newcommand{\fig}[1]{Figure \ref{#1}}
\newcommand{\Fig}[1]{Figure \ref{#1}}

\newcommand{\sect}[1]{Section \ref{#1}}
\newcommand{\Sect}[1]{Section \ref{#1}}

\newcommand{\Alg}[1]{Algorithm \ref{#1}}
\newcommand{\Table}[1]{Table \ref{#1}}
\newcommand{\ws}{\phantom{x}\phantom{x}}
\definecolor{newcolor}{rgb}{.8,.349,.1}

\usepackage[breaklinks=true,bookmarks=false]{hyperref}

\cvprfinalcopy % *** Uncomment this line for the final submission

\def\cvprPaperID{****} % *** Enter the CVPR Paper ID here

\begin{document}

%%%%%%%%% TITLE
\title{Generative One-Class Models\\for Text-based Person Retrieval in Forensic Applications}

\author{David Ger\'onimo and Hedvig Kjellstr\"om\\
Computer Vision and Active Perception Lab, Centre for Autonomous Systems\\
KTH Royal Institute of Technology, Stockholm, Sweden
}

\maketitle
%\thispagestyle{empty}

%%%%%%%%% ABSTRACT
\begin{abstract}
Automatic forensic image analysis assists criminal investigation experts in the search for suspicious persons, abnormal behaviors detection and identity matching in images. In this paper we propose a person retrieval system that uses textual queries (\eg, ``black trousers and green shirt'') as descriptions and a one-class generative color model with outlier filtering to represent the images both to train the models and to perform the search. The method is evaluated in terms of its efficiency in fulfilling the needs of a forensic retrieval system: {\em limited annotation}, {\em robustness}, {\em extensibility}, {\em adaptability} and {\em computational cost}. The proposed generative method is compared to a corresponding discriminative approach. Experiments are carried out using a range of queries in three different databases. The experiments show that the two evaluated algorithms provide average retrieval performance and adaptable to new datasets. The proposed generative algorithm has some advantages over the discriminative one, specifically its capability to work with very few training samples and its much lower computational requirements when the number of training examples increases.
\end{abstract}

%%%%%%%%% BODY TEXT

%%%%%%%%%%%%%%%%%%%%%%%%%%%%%%%%%%%%%%%%%%%%%%%%%%%%%%%%%%%%%%%%%%%%%%%%%%%%%%%%%%%%%%%%%%%%%%%%%%%
\section{Introduction}
\label{section:Introduction}

Forensic image analysis refers to the off-line analysis of images and videos aimed at searching for specific persons or events and inferring their relations in order to provide evidence for a criminal investigation. The data is usually surveillance imagery from a network of surveillance cameras. Some example applications of forensic analysis are suspicious behavior detection \cite{Barbara:2008}, person tracking \cite{Liem:2013}, abnormal behavior detection \cite{Popoola:2012} and identity matching \cite{Farenzena:2010,Bazzani:2013,Layne:2012}. Traditionally, the analysis is performed by human experts who examine the videos. However, in the last decades the amount of data to process has increased drastically, making is difficult to have a fast and accurate manual analysis. Accordingly, automating parts of this process would help to enhance and speed up this task. 

\begin{figure*}[ht!]
\label{figure:annotation}
\begin{center}
\includegraphics[width=\textwidth]{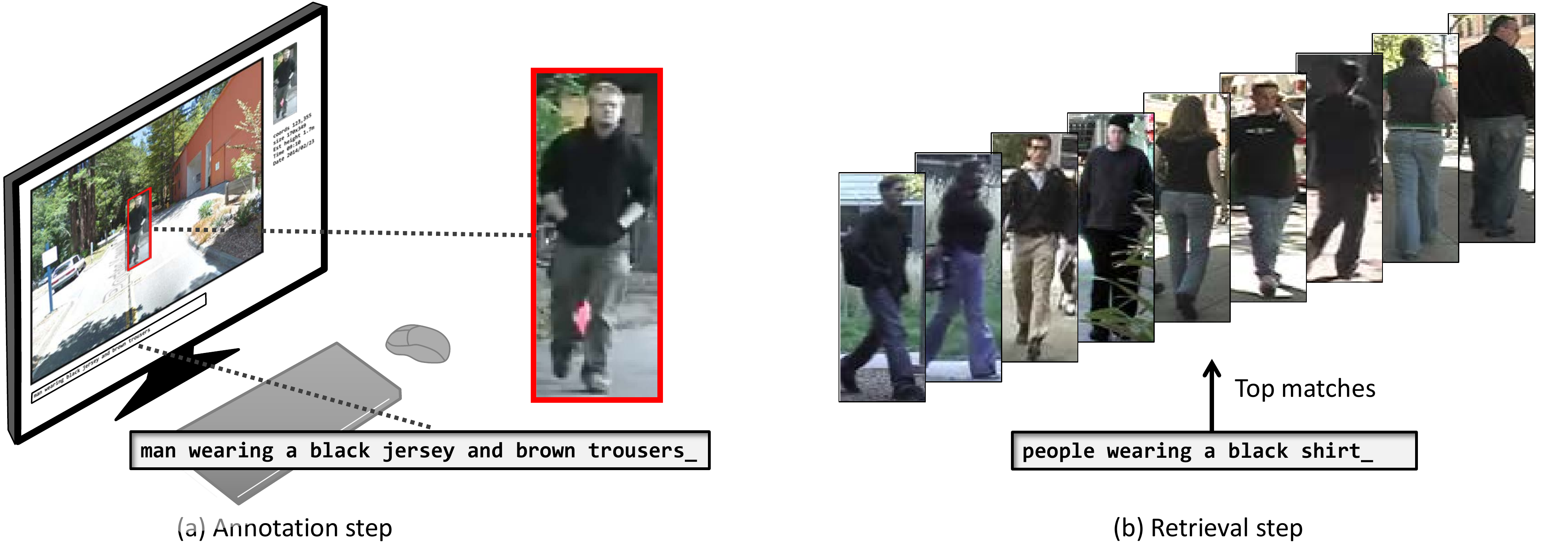}
\caption{Text-based retrieval system. The user annotates and queries the system with textual descriptions of people.}
\end{center}
\end{figure*}

One of the forensic applications that we can envisage is person retrieval based on descriptions provided by witnesses. According to these descriptions, the forensic analyst searches people in a database based on aspects such as gender, height, clothing, etc. One of the most useful searches is using clothing colors, being one of the most distinctive characteristics in a search. The idea is to provide a color-based description like `search a person wearing a blue jacket and black trousers', so the system searches the database retrieving the top matching people according to the description (\Fig{figure:annotation}). From all the components of such a system (\eg, natural language parsing, retrieval visualization, etc.) the most crucial one is the task of defining what the tagged colors are (\eg, `blue' or `black') according to a reduced number of annotations provided by the analyst. In addition from this, one important feature of such a system is the fact that the analyst should be asked only for positive annotations of the (training) samples, not negative ones. This means that the system must be capable of working with annotations referring to single-classes that may overlap with each other, \eg, `blue' and `light'. 

The main contribution of this paper focuses on the task of constructing a robust model from few annotations (which are also input as text descriptions) and using them in the retrieval task. The key requirements of such a system can be divided into the following points:

\begin{itemize}
\item {\em Limited annotation} refers to the capability of working with reduced number of training annotations. 
\item {\em Robustness} is the ability of the system to retrieve relevant results given a query and a trained model.
\item {\em Extensibility} is the possibility of training and searching for specific or generic colors (\eg, `pale beige' or `light') not included in the original model.
\item {\em Adaptability} is the capability of the system to adapt to different conditions (environment, cameras, etc.) given a trained model.
\item {\em Computational cost} under certain ranges both in the training and retrieving steps. 
\end{itemize}

In this paper we propose a system based on a one-class generative model to perform person retrieval based on textual color descriptions. The model consists in a mixture of Gaussians with outlier removal. In order to assess the potential of the proposed generative model in such a forensic application, we compare it to an equivalent discriminative approach (a SVM with a RBF kernel) under the same dataset and conditions. 

The remainder of the paper is as follows. \Sect{section:RelatedWork} reviews the related work. \Sect{section:CommonSystem} describes the textual-query processing and the feature extraction. \Sect{section:Algorithms} introduces the proposed generative model and the compared discriminative one. Experimental results are presented in \sect{section:ExperimentalResults}. Finally, \sect{section:conclusions} summarizes the conclusions.

%%%%%%%%%%%%%%%%%%%%%%%%%%%%%%%%%%%%%%%%%%%%%%%%%%%%%%%%%%%%%%%%%%%%%%%%%%%%%%%%%%%%%%%%%%%%%%%%%%%
\section{Related Work}
\label{section:RelatedWork}
This section reviews the related work in forensic image retrieval and in one-class models in computer vision.

\cite{Vaquero:2009} propose a GUI-based system that performs in-door full-image people search. It uses a face detector, face attribute classifiers (\eg, sunglasses, beard, bald, etc.) represented with Haar-like features, and a torso / legs color classifier based on 8-bin partitions of HSL color space. \cite{Thornton:2011} present another GUI-based system that uses gender, torso/legs color and bags presence in an airport gate context. A generative algorithm representing each body part as a histogram over 12 fixed color names (pre-defined as HSV distribution) is used. Body parts color distributions and torso/legs/head positions are learned through maximum likelihood estimation on a hierarchical model. Contrary to these two approaches, we avoid the use of prefixed and disjoint colors. \cite{Satta:2012,Satta:2014} propose a database retrieval system based on 8 torso and 5 legs text-based color queries (plus two short upper/bottom clothing). The system first clusters a set of random patches based on their color similarity, which are re-clustered to form prototypes. Then, a two-class SVM is trained between positive and negative samples of a given query using multiple component dissimilarity (MCD), \ie, a vector that computes the difference between random image patches and the prototypes. Our proposal differs \cite{Satta:2014} since we avoid fixing the query colors in advance, provide a free-text interface and focus on limited annotations. Moreover, we do not compare with this approach given that it is incompatible to two of the requirements in \Sect{section:Introduction}: extensibility (the colors are fixed a priori and cannot overlap) and limited annotation (the prototypes require all the colors to be annotated in order to retrieve by one color). Finally, it is worth mentioning the research area of garment description for fashion \cite{Liu:2014}. In this case, even though color is often used to describe garments, this problem is very different in several aspects, \eg, number of available annotations or images resolution, so the methods cannot be applied in our case.

	Two early papers working on one-class models are presented by \cite{Moya:1996} and \cite{Ritter:1997}. In the former, a neural network uses multiple objective criteria to model boundaries of the positive class using hyperellipsoids. In the latter, an heuristic method for mixture models estimation parametrizes the outliers within the positive class. \cite{Tax:2001} presents an extensive study on one-class classification methods, in which different approaches to the problem are analyzed, concluding that Parzen density, which is presented as an extension of Gaussian Mixture Models, is the better approach in applications where the training data is close to the test data and the number of training samples is high. Following this idea, we use Gaussian Mixture Models, which is much faster in testing time. Finally, it is worth noting that one-class models have been successfully used in applications such as facial analysis \cite{Zeng:2006} and road detection \cite{Alvarez:2013}.

%%%%%%%%%%%%%%%%%%%%%%%%%%%%%%%%%%%%%%%%%%%%%%%%%%%%%%%%%%%%%%%%%%%%%%%%%%%%%%%%%%%%%%%%%%%%%%%%%%%
%%%%%%%%%%%%%%%%%%%%%%%%%%%%%%%%%%%%%%%%%%%%%%%%%%%%%%%%%%%%%%%%%%%%%%%%%%%%%%%%%%%%%%%%%%%%%%%%%%%
\section{Textual-Query Processing and Feature Extraction}
\label{section:CommonSystem}

The pipeline of the system, both during training and retrieval, is as follows. First, a textual-based front-end processes user queries into color and clothing tuples. Then, pedestrians are segmented from the background and their coarse body parts positions (head, torso, legs) are computed. Finally, the pixels contained in the extracted parts are used as features, which are fed to the classifier either for training a model or using it for retrieval. In this section we overview the two components of the system that are common to the two evaluated classifiers: textual-query processing and feature extraction.

%%%%%%%%%%%%%%%%%%%%%%%%%%%%%%%%%%%%%%%%%%%%%%%%%%%%%%%%%%%%%%%%%%%%%%%%%%%%%%%%%%%%%%%%%%%%%%%%%%%
\subsection{Textual-Query Processing}
\label{subsection:Textual-Query}
Our system uses unconstrained queries in contrast to pre-defined descriptions in \cite{Satta:2014}. Specifically, we use the Stanford Parser \cite{Klein:2003}, an unlexicalized probabilistic context-free grammar (PCFG). It first computes all the possible semantic trees corresponding to the query, which are then assigned a probability that corresponds to the product of the probabilities of each node to split into its children (\eg, a noun phrase is split into determiner and noun nodes with probability 0.3 and into adjectives and noun nodes with probability 0.1). The tree with maximum probability is selected as output and traversed to retrieve the components of the query: each verb phrase is traversed for verbs (\ie, actions), each noun phrase is traversed for adjectives (\eg, colors) and noun entities (\eg, clothing), and in all the cases it searches for conditions (\eg, and, or).

\subsection{Feature Extraction}
\label{subsection:featureExtraction}
The textual-query processing assigns a color label to a specific clothing region (upper or lower garment). In order to extract the features from this region the system segments the pedestrian from the background and localizes its parts. Here we propose a fast and accurate method but any algorithm providing clean shapes will fit. First, the dominant color of a pre-defined rectangular areas in the top and bottom regions is computed using K-means and selecting the cluster with more support. Then, the pixels of the dominant colors in the two regions are used as seeds to perform region growing using GrowCuts \cite{Vezhnevets:2005}. It uses cellular automata to expand the seeds through the image according to the neighbor pixel colors. 

Once the foreground mask is extracted, the position of torso and legs are localized using the approach proposed in SDALF \cite{Bazzani:2013,Farenzena:2010}, which uses color similarity / dissimilarity between horizontal and vertical regions to extract the coarse body parts. It can be seen as an optimization process that tries to find the axes of horizontal symmetry (\ie, the uniform color of a shirt) and vertical asymmetry (\eg, the different colors of a shirt and trousers). 

Finally, we use the HSV color space to represent the pixels. HSV provides a clear separation between colors and lightness in the different dimensions. Experiments using RGB result in poor retrieval performance, probably given that the separation between colors is not clear, \eg, white is a combination of red, green and blue. Experiments using Lab also performed worse than HSV even though Lab uses an independent dimension to represent lightness and two to represent opponent colors.

The notation used hereon is the following. We define as $\mathcal{R}=\{\mathbf{R}_1,\mathbf{R}_2,...,\mathbf{R}_K\}$ the set of samples with a part labeled with a specific color $\mathcal{C}$. Each sample $\mathbf{R}_i$ contains a vector of pixels $\{\mathbf{p}_1,\ldots,\mathbf{p}_n\}$ in HSV space where $\mathbf{p_i}=(p_i^H,p_i^S,p_i^V)$, $p_i^H\in[0^\circ,2\pi]$ and $p_i^S,p_i^V=[0,100]$.

\begin{figure}[ht!]
\begin{center}
\includegraphics[width=\columnwidth]{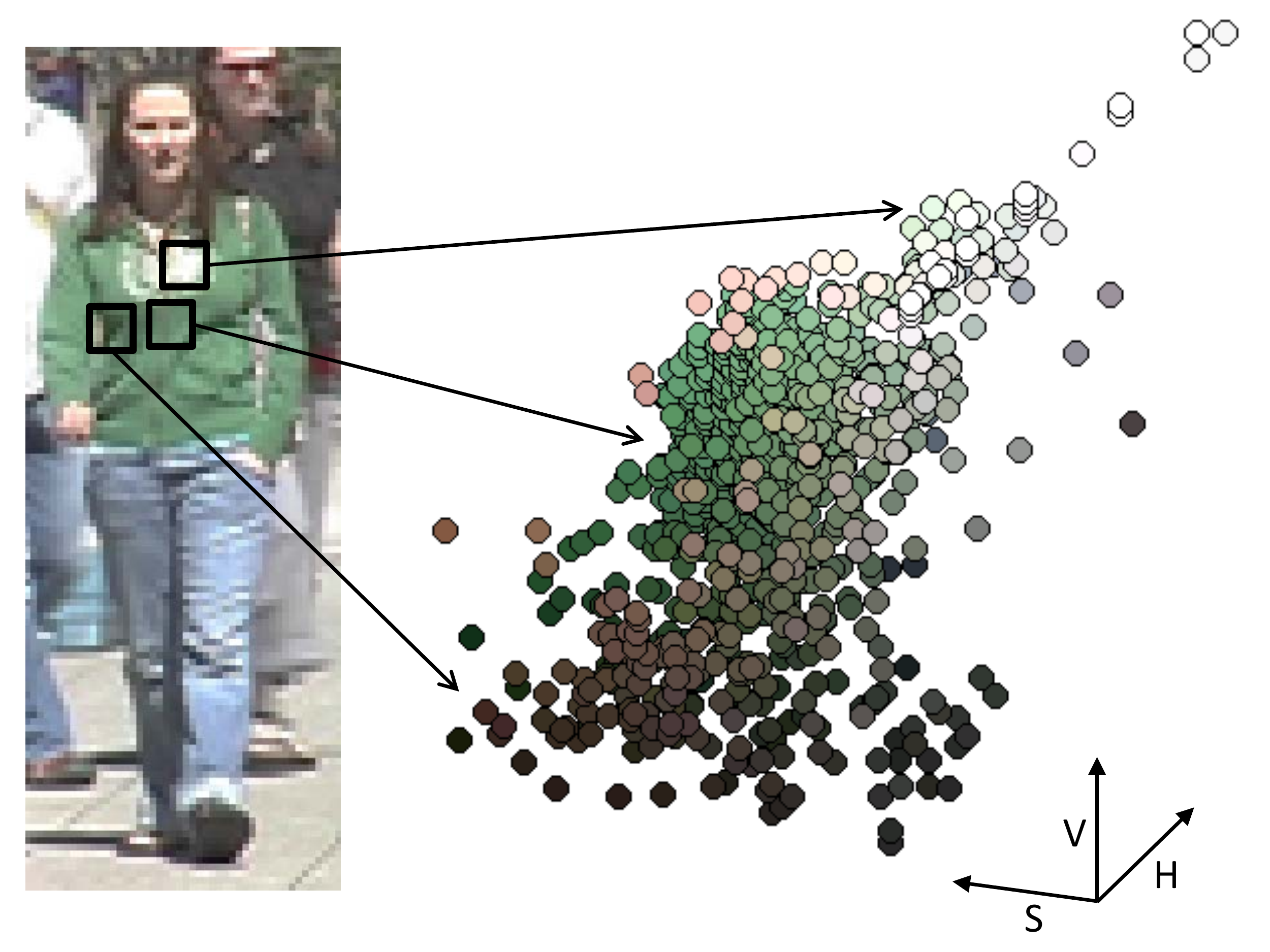}
\caption{Color distribution of the segmented pixels of the jersey}
\label{figure:distribution}
\end{center}
\end{figure} 

%%%%%%%%%%%%%%%%%%%%%%%%%%%%%%%%%%%%%%%%%%%%%%%%%%%%%%%%%%%%%%%%%%%%%%%%%%%%%%%%%%%%%%%%%%%%%%%%%%%
%%%%%%%%%%%%%%%%%%%%%%%%%%%%%%%%%%%%%%%%%%%%%%%%%%%%%%%%%%%%%%%%%%%%%%%%%%%%%%%%%%%%%%%%%%%%%%%%%%%
\section{Evaluated Classifiers}
\label{section:Algorithms}

There are two main paradigms in which to represent a classifier: generative and discriminative models. Generative models  maintain a model of how data was generated given their class. In practice, training a generative model means modeling the underlying distribution of training data given class. The advantage is that comparatively little training data is required, since the imposed distribution model helps interpolate between training examples. However, the disadvantage is that the model is only an approximation of the true distribution. Examples of generative classifiers are Naive Bayes and Gaussian Mixture Models (GMM). 

\begin{figure*}
\begin{center}
\includegraphics[width=\textwidth]{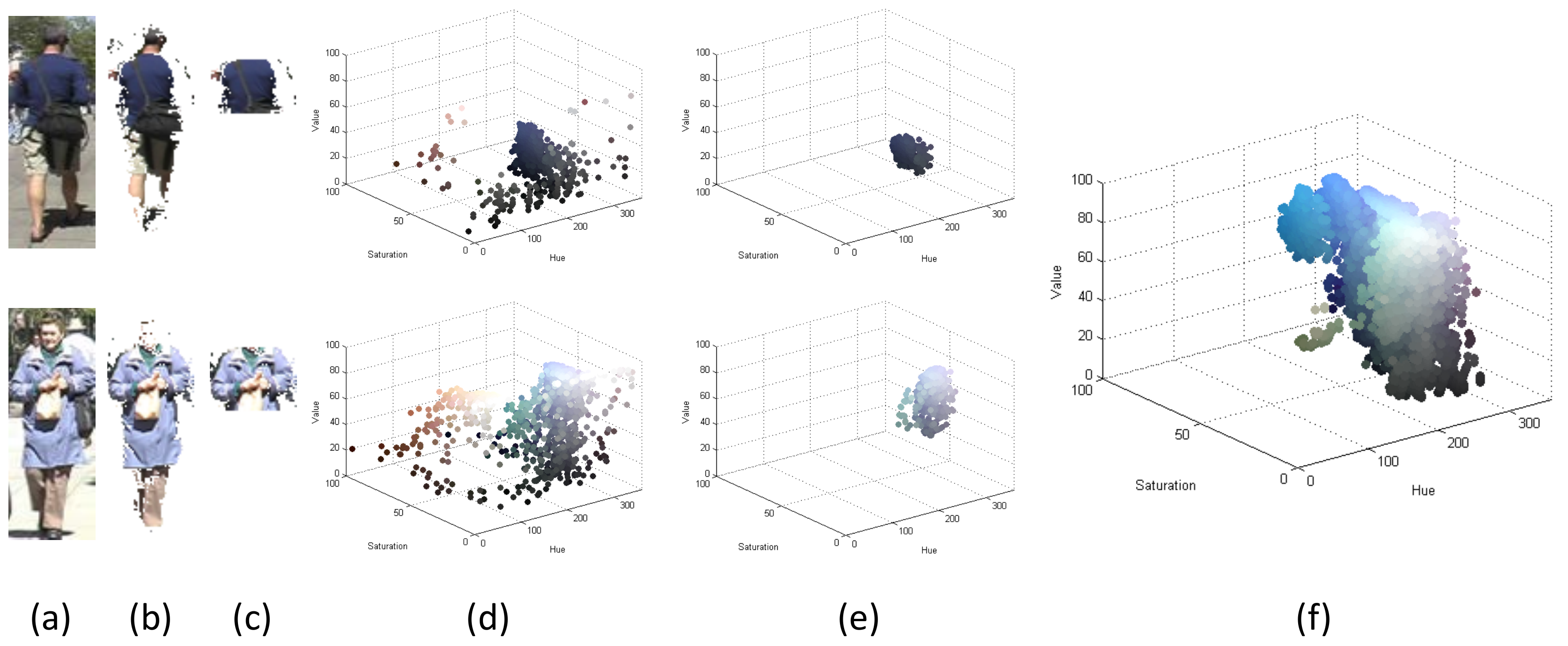}
\caption{Fitlering Process. $(a)$ Original {\em blue} upper clothing samples. $(b,c)$ Foreground segmentation and upper/lower division. $(d)$ Raw HSV pixels. $(e)$ Filtered cloud of each sample. $(f)$ Accumulated $blue$ pixels.}
\label{figure:method}
\end{center}
\end{figure*}

On the other hand, discriminative models are agnostic to how data was generated, and focus on modeling the mapping from observables to the underlying class variable. This has the advantage that no explicit model has to be imposed on the distribution of data given class, which is good in cases where this mapping is very complex. However, discriminative approaches require in general more training data than generative approaches. Examples of discriminative classification approaches are k Nearest Neighbors (kNN), Support Vector Machines (SVM) \cite{Cortes:1995}, and Deep Neural Networks (DNN) \cite{Krizhevsky:2012}.

The specifics of our problem are that we have quite sparse training data with noise and outliers, and that the distribution of data with respect to class are quite ӮiceҬ \ie, representable with an analytic form. This leads us to propose a generative approach, GMM. We compare the chosen approach to a standard discriminative baseline with a similar Gaussian assumption about the data, SVM with Radial Basis Function (RBF) kernel.

%%%%%%%%%%%%%%%%%%%%%%%%%%%%%%%%%%%%%%%%%%%%%%%%%%%%%%%%%%%%%%%%%%%%%%%%%%%%%%%%%%%%%%%%%%%%%%%%%%%
\subsection{Generative Approach: One-Class Gaussian Mixture Model with Outlier Filtering}
\label{sect:oneclasscolormodel}
We propose a generative approach that only uses positive training samples. This aspect is only achievable with this paradigm, which has specific advantages compared to the discriminative one that are analyzed later. The motivation of our algorithm is as follows. As can be noticed in \fig{figure:distribution}, the segmented parts usually contain additional colors to the annotated one. This happens not only because the segmentation can be noisy and contain background pixels but also the clothing can have more colors apart from the dominant one. Given that the proposed generative approach does not have negative examples to correct this, we propose an outlier filtering strategy embedded in the learning.

An intuitive solution to this problem is to accumulate all the pixels from the samples described with a given color and then remove the regions with lower pixel density in the HSV space. However, this process may remove important information from the less represented shades of a color, even if the segmentation is perfect and hue is consistent. In order to amend this, we perform the outlier filtering just after the segmentation and then compute the model with the filtered samples. Hence, the pixel cloud is cleaner from outliers and the less frequent tones also impact the model construction (see \fig{figure:method}). Finally, the color distribution is represented as a Gaussian mixture model of $C$ components. \Alg{algorithm:filtering} details the process.

\begin{algorithm*}[b!]
\normalsize
\label{algorithm:filtering}
\caption{Outlier Filtering Color Mixture Estimation}
\textbf{Input:}\\
{\ws}$\mathcal{R}$: samples with a clothing labeled with a color $\mathcal{C}$.\\
{\ws}$\tau$: filtering threshold.\\
\textbf{Output:}\\
{\ws}$\mathcal{F}$: set of samples filtered of outliers.\\
\textbf{Method:}
\begin{algorithmic}[]
\FOR{$\mathbf{R}_k$ in $\mathcal{R}$}
\STATE Compute the color mean and covariance of the segmented clothing pixels in $\mathbf{R}_k$:
\begin{equation}
\label{equation:mu}
\begin{split}
\hspace{1cm}\mu_{k}^{HSV}=\arctan\left(\frac{\sum_{i=0}^{n}\frac{1}{n}\sin(p^H_i)}{\sum_{i=0}^{n}\frac{1}{n}\cos(p^H_i)}\right)\mathrm{mod}~\pi,~\sum_{i=1}^{n}\frac{1}{n}p^S_i,~\sum_{i=1}^{n}\frac{1}{n}p^V_i{\enspace.}
\end{split}
\end{equation}
\begin{equation}	
\label{equation:covariance}
\hspace{1cm}\Sigma_{a,b}=\frac{1}{n}\sum_{i=1}^{n}\mathrm{D}^a(p_i, \mu_{k})\mathrm{D}^b(p_i, \mu_{k})\forall\,\,\,a,b \in \{H,S,V\}{\enspace.}
\end{equation}
\STATE Filter the outliers of the distribution using Mahalanobis distance:
\begin{equation}
\hspace{1cm}\mathbf{F}_k = \left\lbrace p_i \in \mathbf{R}_k\, \middle|\,\mathrm{Mahal}(p_i,\mu_k;\Sigma)\,<\,\tau \right\rbrace{\enspace,}\hspace{1.2cm}
\end{equation}
\ENDFOR
\STATE Use $\mathcal{F}=\mathbf{F}_1{\cup}\cdots{\cup}\mathbf{F}_K$ to compute a Gaussian mixture model using any estimator (we used EM in our experiments) using Eq. \ref{equation:mu} and \ref{equation:covariance}. 
\end{algorithmic}
\end{algorithm*}

It is important to take into account the circular nature of the hue channel when computing the color distributions, the likelihoods and the mixture estimation. The distance function between two HSV pixels $(\mathbf{p},\mathbf{q})$ is defined as:

\begin{equation}
\begin{split}
\mathrm{D}^{HSV}(\mathbf{p},\mathbf{q}) = \bigg\{&(p^H-q^H+\pi)~\mathrm{mod}\,\,2\pi\,\,-\,\,\pi,\\
 & p^S-q^S,\\
 & p^V-q^V\bigg\}{\enspace}.
\end{split}
\end{equation}

A new image can now be compared to a trained model $\theta_{color}$ to compute its matching to a given query. The log-likelihood of an individual pixel $\mathbf{p}$ with respect to the model $\theta_{color}$ is 

\begin{multline}
\mathcal{L}(\mathbf{p};\theta_{color}) = \\
\sum_{c=1}^{C}\phi_c\mathrm{log}\left(\frac{1}{\sqrt{(2\pi)^3|\Sigma_c|}}\exp{\left(-\frac{1}{2}\mathrm{Mahal}(\mathbf{p},\mu_c;\Sigma_c)^2\right)}\right)
\end{multline}

{\noindent}where $\theta_{color}=\{\phi_1,\mu_1,\Sigma_1,\ldots,\phi_C,\mu_C,\Sigma_C$\} is the mixture model of $C$ Gaussians, being $\phi_i$, $\mu_i$ and $\Sigma_i$ the weight, mean and covariance matrix of each component $i\,{\in}\,C$, and 
\vspace{0.3cm}
\begin{equation}
\mathrm{Mahal}(\mathbf{p},\mathbf{q};\Sigma) = \sqrt{\mathrm{D^{HSV}}(\mathbf{p},\mathbf{q})^T{\Sigma}^{-1}\mathrm{D^{HSV}}(\mathbf{p},\mathbf{q})}{\enspace.}
\vspace{0.4cm}
\end{equation}

{\noindent}\Alg{algorithm:filtering} summarizes the whole filtering process.

{\noindent}The probability that a test sample matches the query is the mean of the pixels likelihood to belong to the $\theta_{color}$ model corresponding to the specified color: 

\begin{equation}
\mathcal{P}(\mathbf{R_{i}};\theta_{color})=\frac{1}{n}\sum_{\mathbf{p}\in\mathbf{R_i}}\exp(\mathcal{L}(\mathbf{p};\theta_{color})){\enspace,}
\vspace{0.4cm}
\end{equation}

{\noindent}where $n$ is the number of pixels in the segmented sample part $\mathbf{R}_i$. 

Extending the single clothing query to multiple clothing using {\em and/or} conditions is then be straightforward: in the case of {\em and} the log-likelihood of the individual clothings is summed, while for {\em or} the score corresponds to the max of the log-likelihood. 

\vspace{0.3cm}

%%%%%%%%%%%%%%%%%%%%%%%%%%%%%%%%%%%%%%%%%%%%%%%%%%%%%%%%%%%%%%%%%%%%%%%%%%%%%%%%%%%%%%%%%%%%%%%%%%%
\subsection{Discriminative Approach: Support Vector Machines with Radial Basis Function Kernel}

We compare the proposed algorithm with a standard discriminative algorithm: Support Vector Machines (SVM). SVM formulate the learning as an optimization problem that maximizes the distance of a high-dimensional hyperplane to the nearest training data points of a positive and a negative class. In other words, the algorithm finds the hyperplane that best separates the positive and negative training examples. The definition of {\em better} is related to the distance of the hyperplane to the nearest examples. This distance is called {\em functional margin} while the nearest examples are called support vectors, and are used to parametrize the hyperplane (\ie, they define the model). Given that the examples are not always linearly separable, they are projected into a higher- or even infinite-dimensional space to be separated. During classification, the algorithm computes the distance and position of every new data point with respect to the learned hyperplane. 

In our application, the model is trained with the segmented pixels of samples labeled with a specific color and garment, and random samples not containing the color. During test time, the probability estimates of the SVM classification (\ie, mean probability of the pixels classification) are used as the score to sort the retrieval results.

%%%%%%%%%%%%%%%%%%%%%%%%%%%%%%%%%%%%%%%%%%%%%%%%%%%%%%%%%%%%%%%%%%%%%%%%%%%%%%%%%%%%%%%%%%%%%%%%%%%
\section{Experiments}
\label{section:ExperimentalResults}
In this section the proposed classifier is evaluated under the requirements in \Sect{section:Introduction}. Given that the focus of the paper is to evaluate the proposed generative algorithm compared to an equivalent discriminative approach, the evaluation of the natural language processing, foreground segmentation and feature extraction are out of the scope of this paper. In the next sections we describe the databases used, parameter tuning for each algorithm, and the behavior of each algorithm with respect to the aforementioned requirements, evaluated as follows:

\setlength{\tabcolsep}{5pt}
\renewcommand{\arraystretch}{1.05}
\newcommand{\spc}{\hspace{12pt}}
\begin{table*}[ht!]
\centering
%\footnotesize 
\caption{BEP (Break-even point) and P@N performance of the proposed generative algorithm (in brackets its difference with respect to the discriminative). Cells without value mean that there are not enough positive training examples in a specific class to perform the experiment.\newline\ }
\small
\begin{tabular}{l@{\extracolsep{6pt}}lllll@{\extracolsep{11pt}}l@{\extracolsep{11pt}}l@{\extracolsep{11pt}}l}

\ \ \ Measure $\rightarrow$& \multicolumn{5}{c}{Break-Even Point Performance (BEP)}                      & \hspace{4pt} P@5      & \hspace{2pt} P@10 & \hspace{2pt} P@20 \\
\cline{2-6} \cline{7-7} \cline{8-8} \cline{9-9}
\#TrainPos $\rightarrow$ & \hspace{15pt} 1& \hspace{14pt} 5         & \hspace{13pt} 10        & \hspace{11pt} 20            & \hspace{10pt} 30          & \hspace{8pt} 10       & \hspace{8pt} 10 & \hspace{8pt} 10 \\
\cline{1-1}  \cline{2-6} \cline{7-7} \cline{8-8} \cline{9-9}
{\em red upper}   & \ 39.1(+6.4)  & 55.1(+4.6)  & 57.8(+1.9)  & 65.9(+6.8)  & \spc-         & \ 94(+8)   & \ 88(+3)  & \ 77(+3)\\
{\em blue upper}  & \ 31.1(+12.6) & 28.5(+1.2)  & 35.4(+2.8)  & 37.8(+1.9)  & 38.8(+1.0)    & \ 52(+28)  & \ 49(+16) & \ 41(+10)\\
{\em white upper} & \ 42.1(+1.7)  & 52.4(-3.7)  & 58.4(+0.0)  & 60.8(-1.6)  & 64.9(+0.3)    & \ 98(+4)   & \ 96(+11) & \ 93(+9)\\
{\em black upper} & \ 47.6(-1.8)  & 60.2(-0.6)  & 62.9(+0.0)  & 65.9(+0.2)  & 66.0(-1.5)    & \ 82(-10)  & \ 75(-13) & \ 72(-12)\\
{\em pink upper}  & \ 25.7(+13.4) & 30.0(+9.1)  & 37.6(+8.6)  & \spc-       & \spc-         & \ 54(+22)  & \ 41(+13) & \spc- \\
{\em green upper} & \ 21.9(+6.1)  & 25.5(-4.2)  & 31.9(-6.1)  & 38.0(-20.8) & \spc-         & \ 64(+14)  & \ 54(+8)  & \ 42(-2)\\
{\em brown upper} & \ 26.1(+5.6)  & 30.2(+4.4)  & 36.3(+8.8)  & 38.3(+6.7)  & \spc-         & \ 62(+32)  & \ 46(+16) & \ 43(+12)\\
{\em gray upper}  & \ 21.1(+7.7)  & 26.2(+5.7)  & 30.8(+4.3)  & 31.4(+3.2)  & \spc-         & \ 44(+8)   & \ 44(+8)  & \ 33(+3)\\
{\em blue lower}  & \ 69.4(+5.0)  & 73.5(-4.9)  & 73.5(-5.5)  & 76.8(-4.3)  & 77.8(-3.4)    & \ 94(+2)   & \ 94(+9)  & \ 92(+6)\\
{\em white lower} & \ 33.5(+9.6)  & 37.1(-3.6)  & 48.5(-4.3)  & 48.9(-10.0) & 51.7(-8.1)    & \ 78(-4)   & \ 73(-8)  & \ 68(-9)\\
{\em black lower} & \ 44.1(+5.7)  & 45.7(-8.7)  & 49.5(-6.1)  & 50.2(-7.2)  & 49.7(-10.8)   & \ 100(-24) & \ 91(+9)  & \ 83(+8)\\
{\em gray lower}  & \ 13.0(+2.3)  & 13.8(+4.8)  & 15.0(+1.2)  & 13.4(-4.6)  & \spc-         & \ 6(-10)   & \ 10(-2)  & \ 12(+1)\\
{\em brown lower} & \ 31.5(+11.0) & 39.0(+19.0) & 41.5(+18.0) & \spc-       & \spc-         & \ 44(+22)  & \ 45(+20) & \spc- \\
\cline{1-1}  \cline{2-6} \cline{7-7} \cline{8-8} \cline{9-9}
Mean              & \ 34.3(+6.5)  & 39.7(+1.7)  & 41.7(+1.8)  & 47.9(-2.7)  & 58.1(-4.2)    & \ 67(+7.0) & \ 62(+6.9)& \ 60(+2.6)\\
\end{tabular}
\label{table:BEP}
\end{table*}

%-----------------
%                 1 ex      5 ex      10 ex         20 ex
%
%red upper    ->  38.3      52.1                    64.8
%blue upper   ->  18.0      28.8                    34.2
%white upper  ->  40.8      52.5                    62.2
%black upper  ->  41.6      57.9                    63.0
%pink upper   ->  15.7      27.6                    32.3 x
%green upper  ->  24.7  		26.9										35.0
%brown upper  ->  30.2      30.0                    42.2
%gray upper   ->  17.1      26.2                    29.4
%
%blue lower  ->   60.2      73.4                    75.9
%white lower  ->  26.9			33.3										43.9
%black lower  ->  41.0			48.5 										54.4
%gray lower   ->  10.0			11.9									  10.7
%brown lower  ->  12.0			25.5									  31.0 x
%-----------------                                  
%mean             28.96      36.5                    46.8

\begin{itemize}[noitemsep]
\item {\em Limited annotations}. All the experiments are performed with limited and increasing number of training examples.
\item {\em Robustness}. A set of basic queries are evaluated (\Sect{subsection:robustness}), extracting two measures: Break-Error Point (BEP), which computes the crossing point between the precision-recall curve ($\text{precision}=\frac{TP}{TP+FP}$, $\text{recall}=\frac{TP}{TP+FN}$), and  at N (P@N), which measures the precision in the first top-N results. 
\item {\em Extensibility}. We analyze the required effort and the capability of the system to incorporate colors and new samples of a given color in \Sect{subsection:extensibility}.
\item {\em Adaptability}. Cross-database tests emulate a data corpus acquired in multiple conditions and test the capacity of the approaches to generalize (\Sect{subsection:adaptability}).
\item {\em Computational cost}. In \Sect{subsection:cost}, we analyze the cost of training and testing of the two evaluated algorithms. 
\end{itemize}

\subsection{Databases}
We perform the main experiments using the VIPER re-identification database \cite{Gray:2007}, which contains 632 pedestrian image pairs (we use all 1264 pedestrians indistinctly) of $128\times48$ pixels, using the color annotations provided by Satta {\etal}\footnote{http://pralab.diee.unica.it/en/AmbientIntelligence} in \cite{Satta:2014}. For the adaptability experiments, we use the PRID \footnote{http://lrs.icg.tugraz.at/datasets/prid} and QMUL \footnote{http://www.eecs.qmul.ac.uk/$\sim$ccloy/downloads} re-identification databases with our own text annotations. We select 1043 and 531 samples from PRID and QMUL, respectively, discarding blurred and occluded pedestrians, and then resize them to $128\times48$ pixels.

It is important to point out that a different foreground segmentation algorithm is used depending on the database. For VIPER we use the masks provided by Bazzani {\etal}\footnote{http://www.lorisbazzani.info/code-datasets/sdalf-descriptor} \cite{Bazzani:2013}, which are computed using SCA (structure element component analysis) \cite{Jojic:2009}, which models an image class as a mixture of segments where pixels of each segment have strong self-similarity.
For PRID and QMUL we use the algorithm introduced in \Sect{subsection:featureExtraction}. The use of different algorithms is fair in the sense that it can be seen as a property of the database like lighting or perspective. 

\subsection{Parameter Tuning}
The parameter tuning of the algorithms prior to the experiments is performed as follows. In the generative algorithm, different settings of $\tau$ (filtering threshold) and M (mixture components) have been tested. In the case of $\tau$, the performance is highest when set to 1 or 2, then it progressively decreases until $\tau=\infty$, which corresponds to using no outlier filtering. The benefits of the filtering also depend on the number of training samples, \ie, as the training samples decrease the proportion between outliers and outliers increases. As an example, the performance improvement thanks to filtering can range from $5\%$ to $1\%$ when using 1 to 20 training samples. Tests with different components ($M=\{1,\cdots,5\}$, fixing $\tau=2$) show that even though average performance is stable, in models with less training samples a small M works better, otherwise the components adjust to specific samples undermining the model generality. On the contrary, in models with a high number of samples, a higher M helps to adjust the model to the singularities of the data. We set $M=2$ as a tradeoff between these two aspects. 

In the discriminative approach three kernels are evaluated using LIBSVM \cite{Libsvm}: linear, radial basis function with different $\gamma$ and polynomials of different degrees. Every parameter is evaluated using 5-fold cross-validation according to a search grid also taking into account the cost $C$. The best configuration is a RBF kernel with $\gamma=0.01$ and $C=1$, overperforming linear and polynomial both in performance and computational time, which leads to conclude that Gaussians are better suited to represent the color data as we hypothesized.

\subsection{Robustness}
\label{subsection:robustness}

We test the algorithms for a set of basic queries corresponding to the class annotations in \cite{Satta:2014}. For each query, the database is first divided into training and testing, each containing the same number of positive examples. Then  each classifier is trained with a number of positives (plus the same number of negatives in the discriminative algorithm) and used to evaluate the full testing set.  There is not a single perfect way of selecting the sample number in both approaches (we either have twice the total examples in the discriminative, or twice the number of positives in the generative), so we opt to fix the number of positives as the fairest approach.

\begin{figure*}[!ht]
\begin{center}
\includegraphics[width=\textwidth]{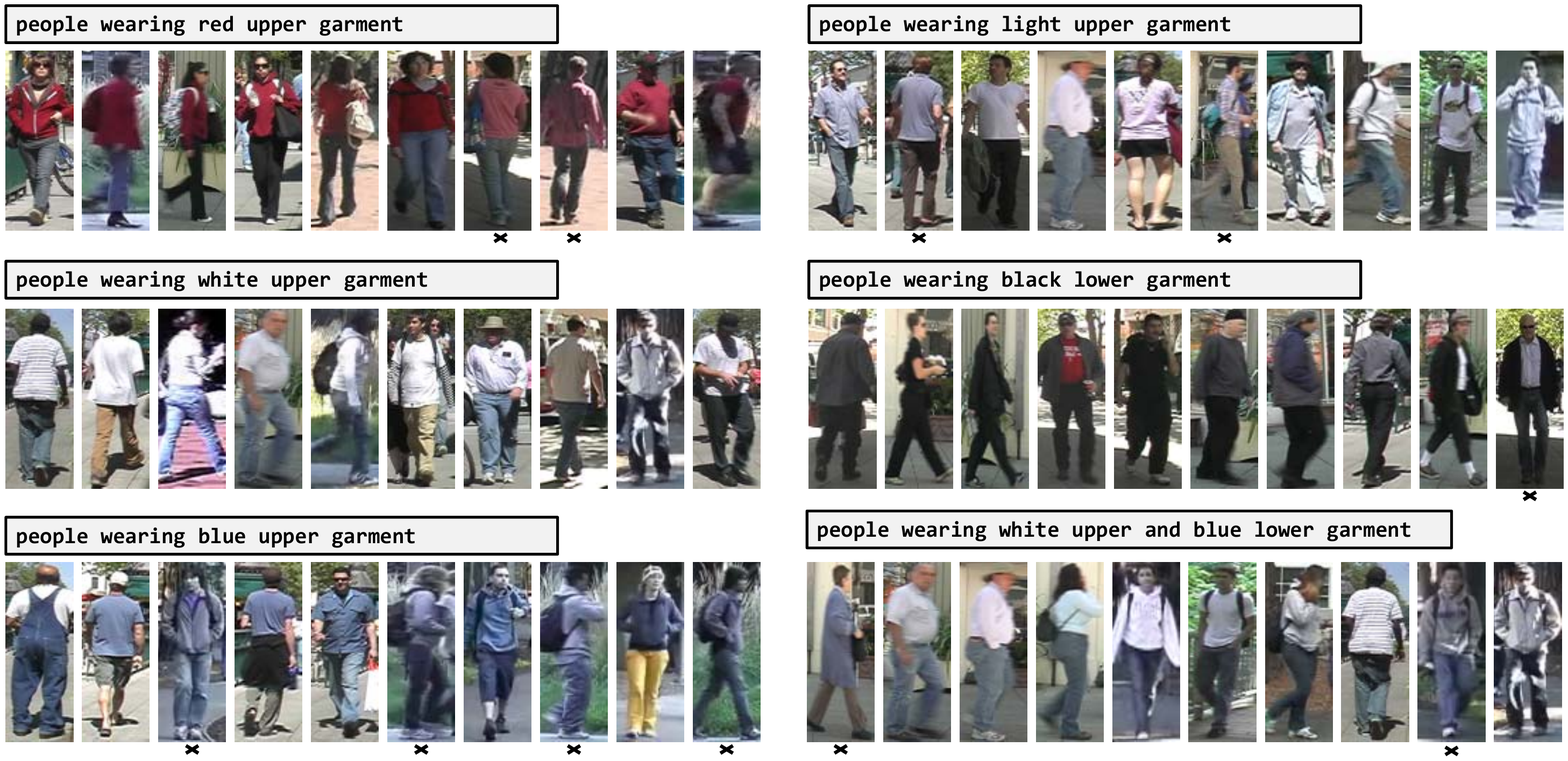}
\caption{Qualitative results. The top-10 results of six different queries using the proposed generative algorithm and 10 training positive examples in VIPER dataset, sorted according to their matching probability. Each X identifies the false positive results for the given query description.}
\label{figure:qualitative}
\end{center}
\end{figure*}

The above process is repeated ten times, providing the results summarized in \Table{table:BEP}. The first five columns list the BEP performance for the generative approach and its difference with respect to the discriminative. Two trends can be noticed. First, increasing the number of positive training examples increases the performance in both approaches. Second, with limited training positives the generative approach has higher performance. This trend inverts when the number of training positives goes from 10 to 20. The last three columns show the P@5, P@10 and P@20 when using 10 training positive examples to build the model. The performance is acceptable for most of the queries, clearly beating the discriminative approach in 8 of them. It is also worth analyzing the cases in which the performance is very different in the two algorithms: in heterogeneous colors (\eg, blue, pink or brown) the generative algorithm seems to model better the space, while homogeneous colors (\eg, black) are better modeled by the discriminative one. 

\Fig{figure:qualitative} illustrates some qualitative results of the proposed approach demonstrating the ability of the proposed system to retrieve garments matching to a text-query search, even if some of the classes overlap (\eg, white and light). 
\vspace{0.5cm}

\subsection{Extensibility}
\label{subsection:extensibility}
\begin{figure}[!h]
\begin{center}
\includegraphics[width=\columnwidth]{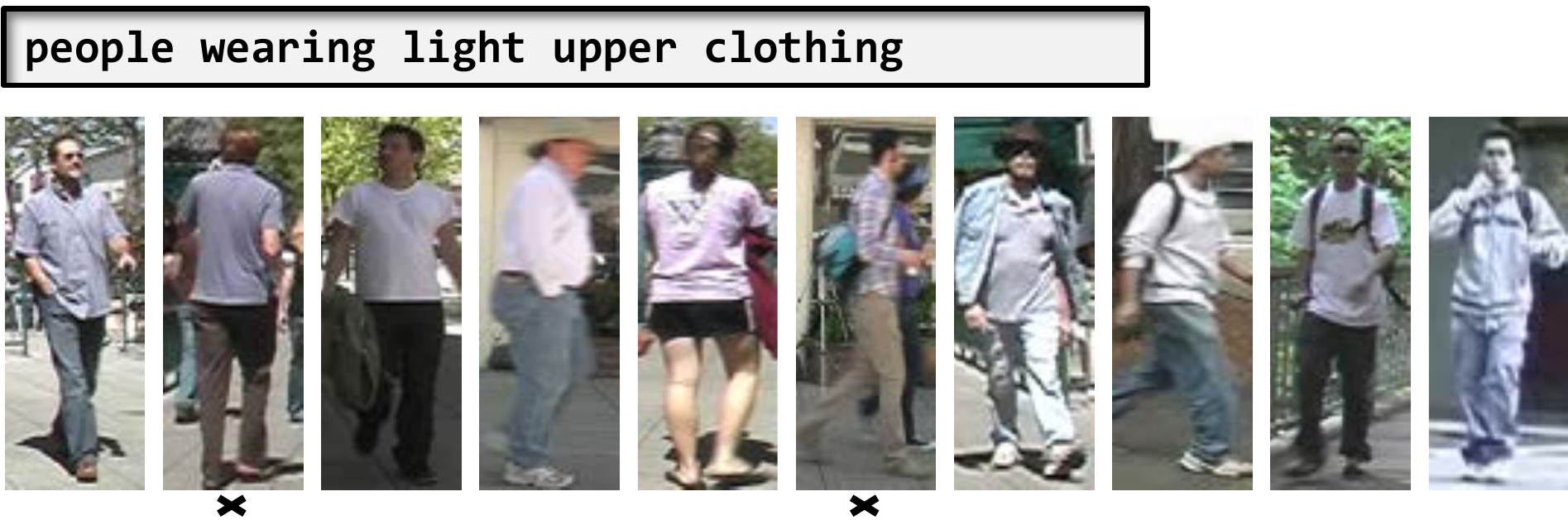}
\caption{Qualitative results demonstrating the system's extensibility. As can be seen, the system is capable of successfully working with other color concepts such as '$light$'.}
\label{figure:qualitative-light}
\end{center}
\end{figure}

In this section we add two queries to the system and analyze its cost and performance: generic {\em light and dark upper garments}. The results are quite similar in both algorithms, even if the generative one is still better. Using a model trained with 10 positive examples, the BEP (averaged from 10 experiments) for {\em light upper} and {\em dark upper} queries is $54.5\%$ and $75.9\%$ in the generative and $52.8\%$ and $74.9\%$ in the discriminative. The P@N is also comparable. \Fig{figure:qualitative-light} illustrates some qualitative results.

The first important advantage of the generative algorithm compared to other related approaches \cite{Satta:2014} is that it does not need to re-train the whole system if a new query is added. This has the disadvantage of not being able to re-use annotations of a given class as negatives for the rest of the classes, but saves from having to re-train everything if a new query is added. The second advantage is that the algorithm can manage overlapped classes, \eg, light and white models/queries in the same system without re-training. This not only benefits the but also allows the forensic analyst to be either specific (\eg, define a specific tone of a color) or generic (\eg, a description grouping several colors in a single illumination).

\subsection{Adaptability}
\label{subsection:adaptability}
The adaptability of the models to new databases is evaluated using cross-database experiments: a model using 10 positive VIPER samples is trained and then used to retrieve in PRID and QMUL. \Table{table:cross} shows the mean BEP of the 6 queries that exist in all databases (average of 10 representative experiments): red, blue and white upper garment, blue, black and brown lower garment. 

\begin{figure}[hb!]
\begin{center}
\includegraphics[width=\columnwidth]{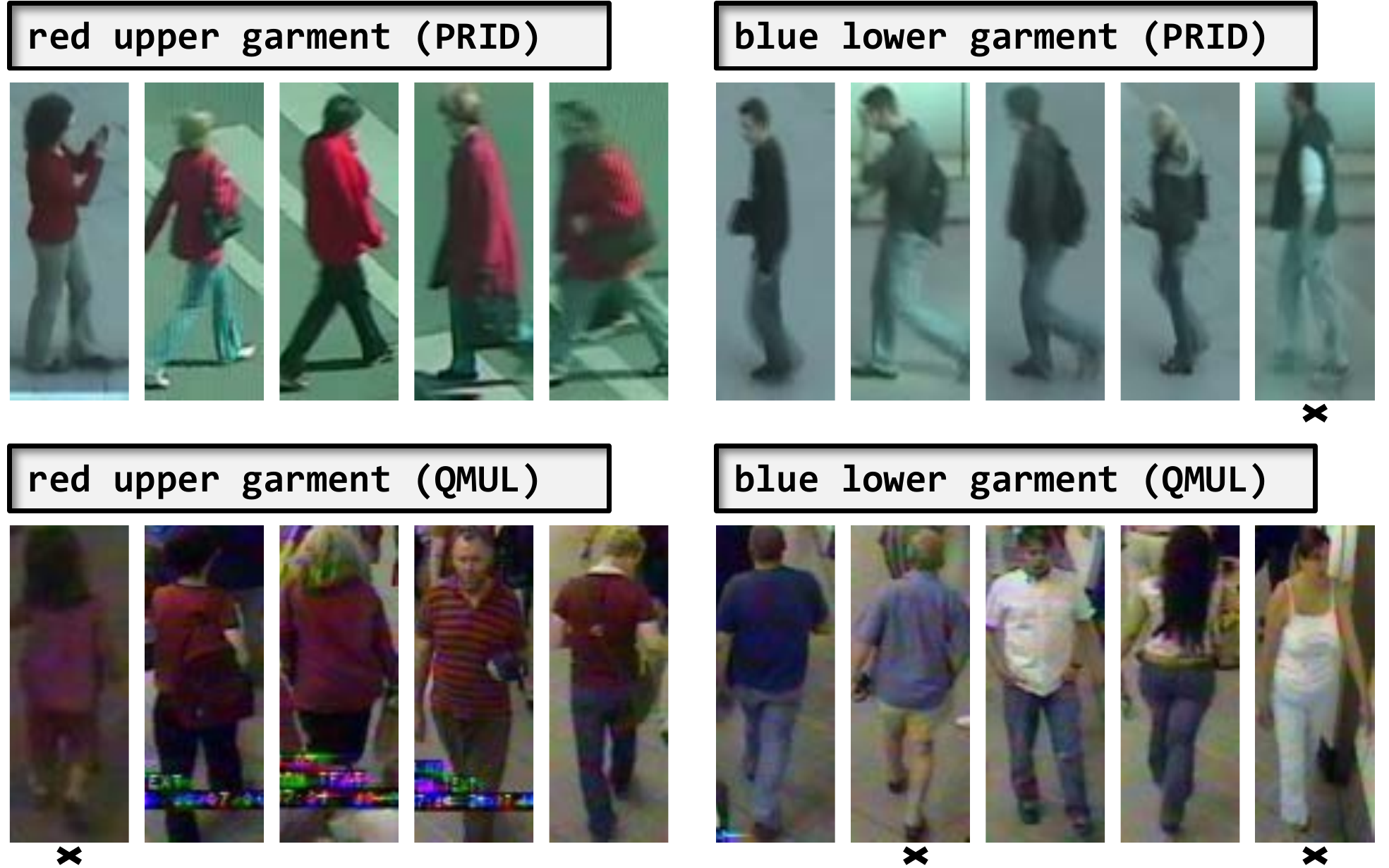}
\caption{Cross-database qualitative results using the generative algorithm. Top-5 retrieved examples for different queries.}
\label{figure:adaptation}
\end{center}
\end{figure}

Two conclusions can be extracted. First, the performance of the generative approach tends to be higher than the discriminative, which means that it generalizes better. Second, as already seen in \Table{table:BEP}, the more limited the number of training samples is, the bigger the difference in performance between the two algorithms. \Fig{figure:adaptation} illustrates some qualitative results. Notice the difference in illumination conditions, background clutter and compression artifacts compared to \Fig{figure:qualitative}.

\setlength{\tabcolsep}{5pt}
\renewcommand{\arraystretch}{1.05}
\begin{table}[h!]
\centering
%\footnotesize 
\caption{Cross-database results. PRID and QMUL retrieval using VIPER model. Generative algorithm performance (difference w.r.t. discriminative in brackets).\newline\ }
\small
\begin{tabular}{l@{\extracolsep{8pt}}l@{\extracolsep{15pt}}l@{\extracolsep{15pt}}l}
         & \multicolumn{3}{c}{Break-Even Point Performance (BEP)} \\
\cline{2-4}
\#TrainPos $\rightarrow$ & \hspace{17pt}1            & \hspace{17pt}10           & \hspace{17pt}30          \\
\cline{2-2} \cline{3-3} \cline{4-4}
PRID        &  38.8(+8.08) & 33.9(+4.58)  & 34.3(+1.25)       \\
QMUL        &  52.6(+8.48) & 49.71(-1.04) & 50.0(-0.23)       \\
\end{tabular}
\label{table:cross}
\end{table}

\subsection{Computational Cost}
\label{subsection:cost}
\begin{figure}[ht!]
\begin{center}
\includegraphics[trim=3cm 10.2cm 3cm 10.25cm, clip=true, width=1.03\columnwidth]{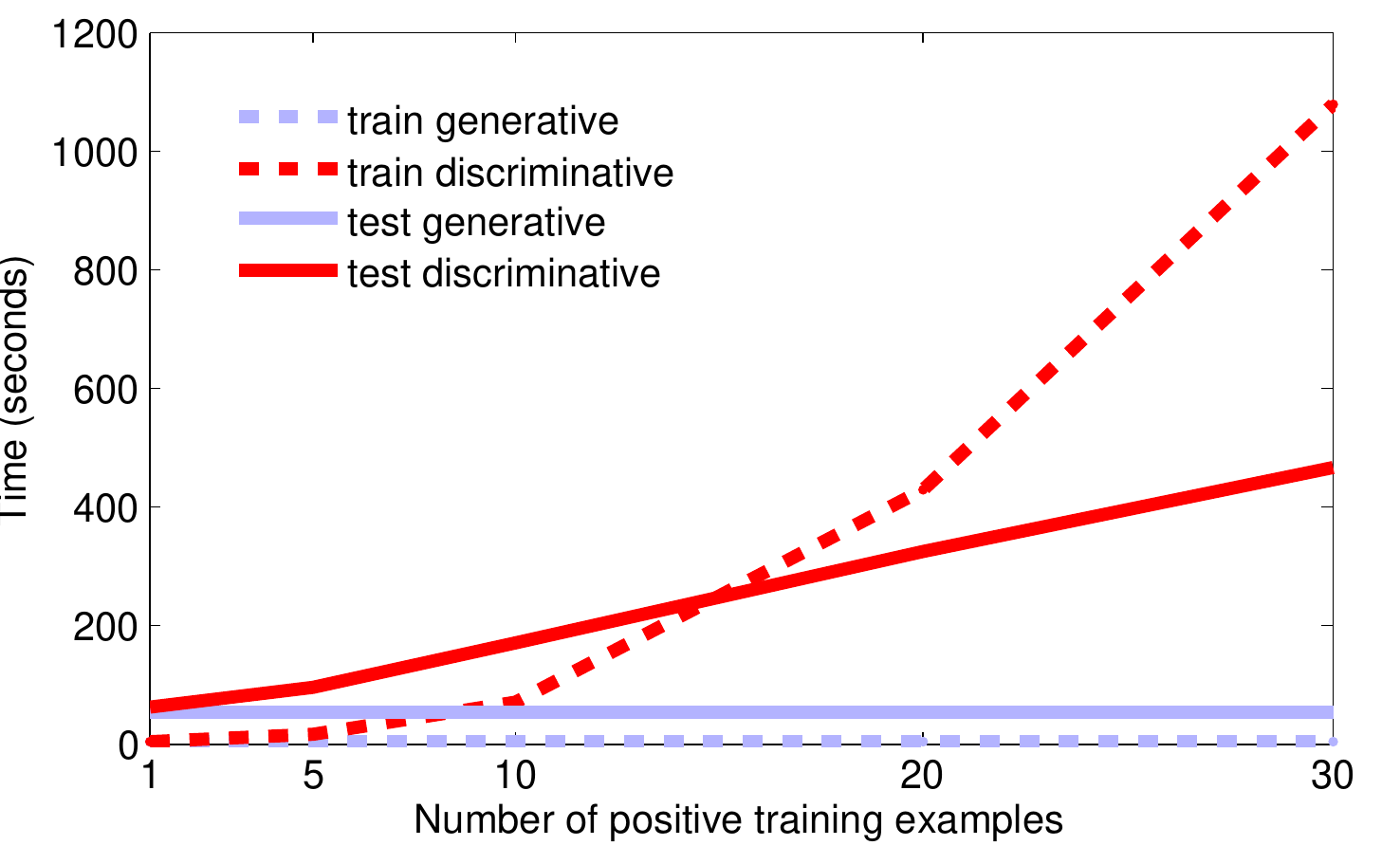}
\caption{Computational cost of each algorithm (average of 10 experiments).}
\label{figure:cost}
\end{center}
\end{figure}

Experiments are performed using models trained with 1, 5, 10, 20 and 30 positive examples, each experiment being repeated 10 times to show the average cost during training and retrieval. All experiments are performed using Matlab. \Fig{figure:cost} illustrates that both approaches have similar training cost when using few examples. However, as this number increases, the computation time for the discriminative one also increases while the generative is very low. In the case of the retrieval time, there is again a direct relation between the number of samples, which affects the complexity of the model, and the computation time. The retrieval time multiplies by 10 for every additional positive training example in the case of the discriminative approach while the generative one is very constant. The explanation to this is that while the SVM may add new support vectors for each new training example, which have to be evaluated when testing, the GMM is parameterized by a constant number of Gaussians independently from the number of samples. 

\vspace{0.5cm}

\section{Conclusions}
\label{section:conclusions}
We have presented a person retrieval system that builds on one-class color models and text-based queries. The proposed generative model with outlier filtering has several advantages for forensic analysis: it can work with a limited number of annotated samples, it adapts well to other databases, it allows to extend the models with new overlapped colors (\eg, ``blue jersey'' and ``light jersey'') without retraining the whole system, and it is less computationally demanding than the discriminative algorithm. 

Some future work directions include a process to feed the model with new samples in a bootstrapping way (to improve the performance in a given database), or domain adaptation approach (to adapt better to new databases), which in our generative model would be straightforward; the combination of generative and discriminative approaches \cite{Bishop:2007}; and the extension of the system to other queries such as gender or textured clothing, inspired in re-identification approaches \cite{Layne:2012}. 

\vspace{1cm}

\section*{Acknowledgements}
This work has been funded by the Beatriu de Pin\'os PostDoctoral Fellowship (EU Marie Curie Actions) of the Generalitat de Catalunya and by the Swedish Research Council under the project 2013-5473 FOVIAL.

{\small
\bibliographystyle{ieee}
\bibliography{template}
}

\end{document}